\documentclass[sigconf,natbib=true]{acmart}
\usepackage{times}
\usepackage{soul}
\usepackage{url}
\usepackage{graphicx}
\usepackage{amsmath}
\usepackage{amsthm}
\usepackage{booktabs}
\usepackage{algorithm}
\usepackage{algorithmic}
\usepackage{float}
\usepackage{bbm}
\renewcommand\footnotetextcopyrightpermission[1]{}
\begin{document}

\title{Toward An Optimal Selection of Dialogue Strategies: A Target-Driven Approach for Intelligent Outbound Robots}





\author{
  Ruifeng Qian$^*$, 
  Shijie Li$^*$, Mengjiao Bao, Huan Chen, Yu Chen \\
  Meituan \\
  \texttt{\{qianruifeng,lishijie11,baomengjiao,chenhuan15,chenyu17\}@Meituan.com} \\
}
\thanks{$^*$ Both authors contributed equally to this research.}


\begin{abstract}
  With the rapid development of the Internet and the vigorous growth of the artificial intelligence technologies, intelligent outbound robots have attracted more and more attention from the industry, especially FinTech companies. Intelligent outbound robots, also known as intelligent agents, are used to replace humans to complete a specific spoken interactive task, such as market survey, sales calls, debt collection, etc. And different intelligent outbound robots have different accomplishment targets respectively. The currently widely used intelligent outbound robots are mainly flow-based methods and behavior-cloning-based methods. However, none of these methods can guarantee that the retrieval of the robot's strategies is effective. To address this challenge, we propose a target-driven framework based on sequence prediction and streaming inference called Policy2Target. The proposed Policy2Target framework defines the dialogue between the agent and the customer as a sequence labeling problem. In the training phase, an attention-based model is designed to optimize a sequence labeling task and establish the mapping of the agent strategy to the accomplishment probability of the target (such as the repayment probability of debt collection in this paper). When making online predictions, the trained model uses the streaming inference method to predict the accomplishment probability at each time step to select the most effective agent strategy. This framework can make full use of a large amount of manual outbound calling data and obtain a better effect for the accomplishment target. Extensive experiments are conducted and demonstrate the effectiveness of the proposed framework. We have also deployed the framework online and observed significant improvement compared to the flow-based method and the behavior-cloning-based method in a rigorous A/B test.
\end{abstract}




\maketitle
\section{Introduction}
With the growth of the economy and society, enterprises, especially in the FinTech industry, have increasing demands of outbound calls for customers such as debt collection, marketing, anti-fraud calls, and so on. But a large amount of repetitive and mechanical work occupies most of the time of human agents, so the cost of equipment and labor for enterprises is increasing accordingly. At the same time, with the development of artificial intelligence technology in the past few decades, it has become quite common for companies to use new technologies such as Big Data and artificial intelligence to empower outbound call businesses.
The intelligent outbound robot is a typical application of the artificial intelligence technology in the field of outbound call businesses. It is mainly used to communicate with customers in order to accomplish a certain target. It has the characteristics of low cost, high reuse, and easy compliance, which has attracted more attention from the industry. 

\begin{figure}[htbp]
\centering
\includegraphics[scale=0.3]{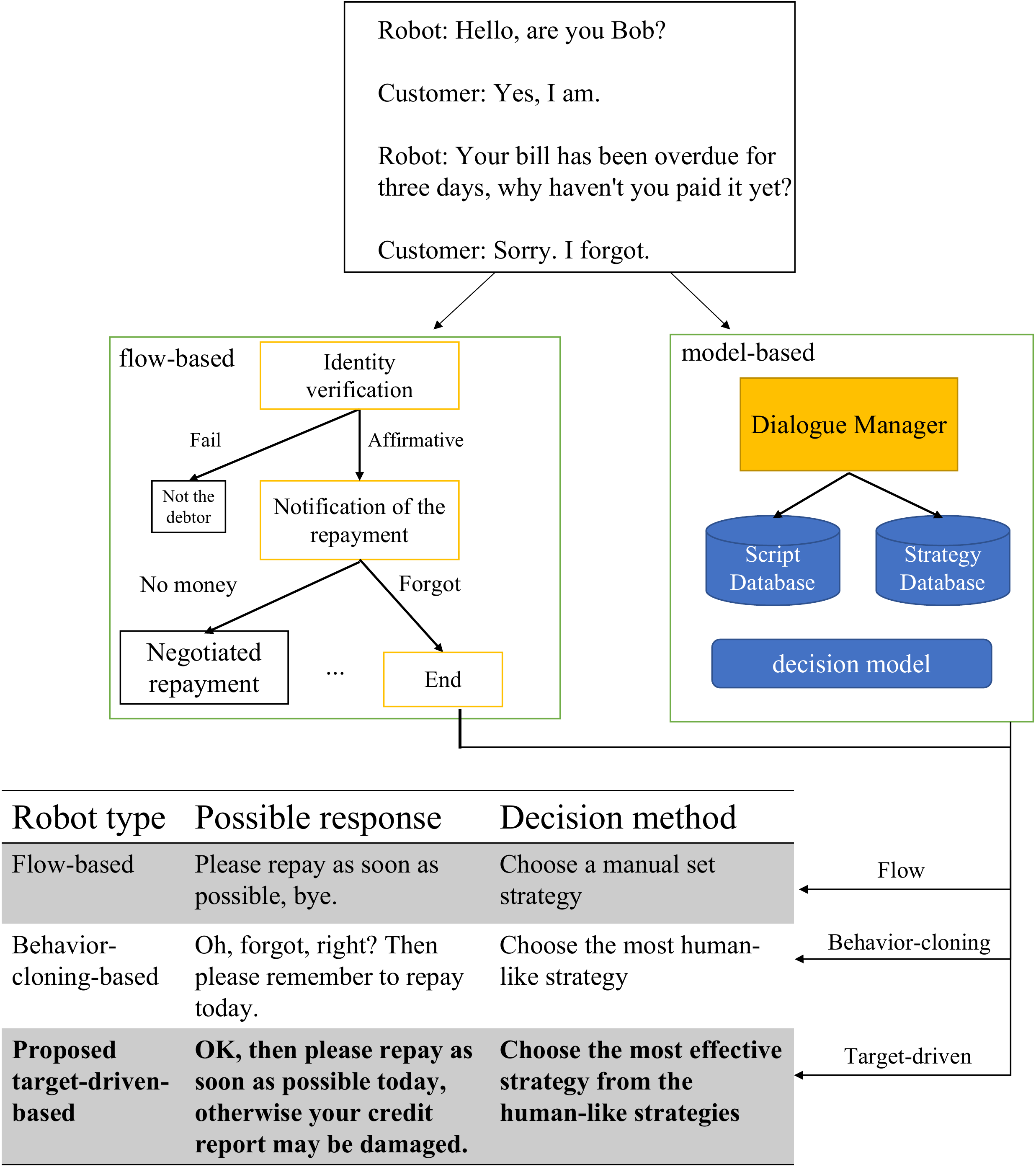}
\caption{Illustration of different frameworks of intelligent outbound robots.}
\label{fig:label}
\end{figure}

At present, there are two kinds of intelligent outbound robots in the industry but both of them still leave large room for improvement. One kind of them is based on a finite state machine relying on the configuration of jump conditions and corresponding nodes based on manual experience. This kind of intelligent outbound robot is also called a flow-based robot. For example, the schematic diagram of the working model of a flow-based robot for debt collection is shown in Fig.\ref{fig:label}. In each round, the robot will reply to the user with the words corresponding to each node. 

Another type of intelligent outbound robots is based on model predictions. It uses the model to imitate the behavior of human agents to select the reply to the current round of dialogue. This type of robot is also called a behavior-cloning-based robot \cite{wang2020two,wang2021ifdds}.

The main drawback of the flow-based model is that it requires a lot of expert experience as the basis for configuration. When the jump relationship between nodes becomes complicated, the configuration becomes particularly cumbersome and the maintenance cost becomes correspondingly high. In addition, the flow-based model uses less contextual information, and the jumped node mainly depends on the result of intent recognition, which requires extremely high accuracy of the intent recognition module.

In comparison, the behavior-cloning-based model mainly relies on learning the actions of the human agents to automatically generate the node jump relationships. This method does not require too much configuration cost, but due to the strong randomness of the actions of the human agents, this may make model training more difficult. In addition, the agent strategy chosen from the behavior-cloning-based model is probably not the most effective strategy since the behavior-cloning-based robot does not evaluate the effectiveness of the strategy but only the anthropomorphism.
\begin{figure}[htbp]
\centering
\includegraphics[scale=0.5]{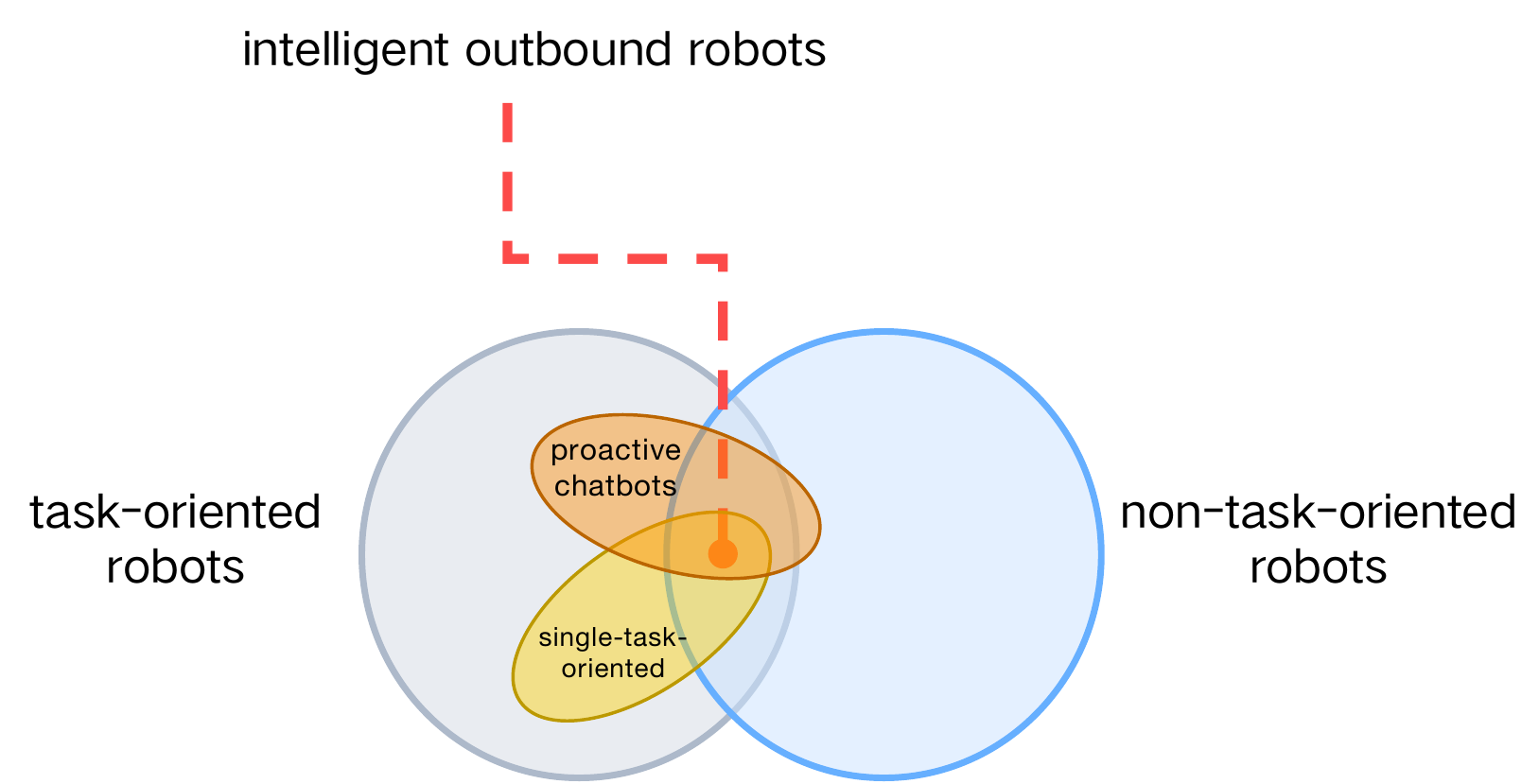}
\caption{The technical classification of intelligent outbound robots.}
\label{fig:outboundrobots-definition}
\end{figure}

In this paper, to tackle these problems, a sequence modeling framework is designed based on target completion probability prediction. The purpose of the framework is to predict the effectiveness of the human strategy, filling the gap between the strategy and the dialogue target for better strategy retrieval illustrated in Fig.\ref{fig:label}. The framework can learn the most effective dialogue strategy while taking into account the advantages of easy configuration of the behavior-cloning-based robots. 
\\
\textbf{Contributions.} We summarize our contributions as follows.

(1) We formally introduce intelligent outbound robots and distinguish them from task-oriented robots and non-task-oriented robots. At the same time, we also formally define the optimization target of an intelligent outbound robot. (Section 2 \& Section 3)

(2) We design and develop a novel framework based on sequence annotation modeling, which can bridge the gap between the dialogue strategy and the dialogue target, solving the problem of the bias between the quality of human agent behavior fitting and the effectiveness of the robot. (Section 3)

(3) An attention-based structure is designed to assist the model to obtain a better-aggregated vector for the strategy based on the user profile. Besides, a two-way multi-task joint learning is proposed to employ the user intent to guide the collection strategy selection. And a novel explicit memory tracker module is designed and applied to make the robot's reply more contextual and logical. (Section 4)

(3) A new evaluation metric WBAUC (Weighted Bins Area Under ROC) is proposed to solve the performance measurement problem under the influence of user bias, and several empirical experiments of our framework Policy2Target are conducted to demonstrate its effectiveness. (Section 5)
\begin{figure*}[ht]
\centering
\includegraphics[scale=0.22]{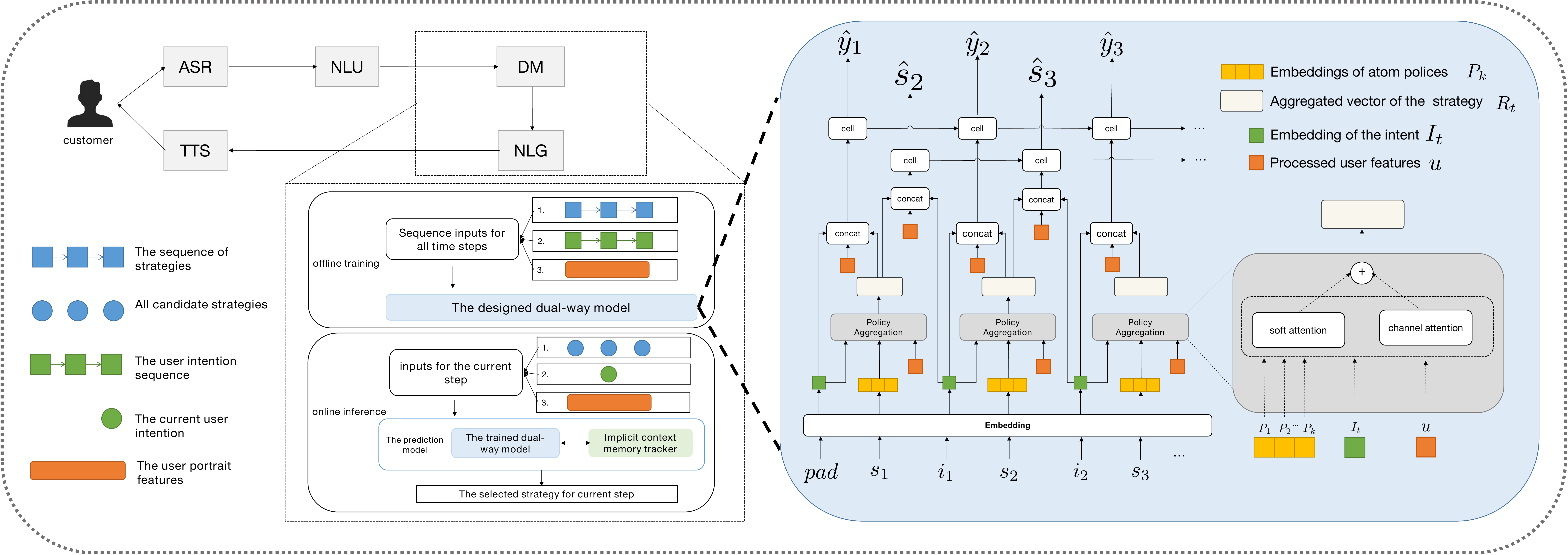}
\caption{Overview of the proposed Policy2Target Framework.}
\label{fig:framework-overview}
\end{figure*}
\section{Related Work}
\textbf{Spoken dialogue systems.} For a long time, spoken dialogue systems have attracted the attention of both academia and industry due to their great academic and commercial value. Typically, dialogue systems can be divided into two different types: (1) task-oriented systems and (2) non-task-oriented systems, also called chatbots \cite{chen2017survey}. A task-oriented dialogue system is generally used to perform a specific task. For example, a debt collection robot is used to complete the task of "collecting overdue debts". Correspondingly, a non-task-oriented system does not complete a specific task, and it only needs to interact with people to provide reasonable responses and some entertainment functions. In particular, for the dialogue system related to this article, there are generally two main solutions: (1) modular-based systems and (2) end-to-end systems \cite{ni2021recent}. A typical modular-based system usually consists of a natural language understanding module (NLU), a dialogue management module (DM, generally composed of a dialogue state tracker and a dialogue policy learning module), and a natural language generation module (NLG) \cite{chen2017survey}. A task-oriented dialog system requires more strict control over the dialog flow since it aims to accurately process and respond to user messages. Modular-based methods can generate responses in a more controllable way, so many industrial solutions are based on modular methods \cite{henderson2014second,kim2018two,hu2020sas}. Of course, some work in academia focus on the end-to-end methods \cite{balakrishnan2019constrained,wang2019incremental,dai2020learning},but they are not the mainstream in the industry.
\\
\textbf{Intelligent outbound robots.} The intelligent outbound robot is a typical application of the task-oriented dialogue system in the industry. However, intelligent outbound robots are different from ordinary task-based robots since the dialogue of the intelligent outbound robots is generally initiated by the robot, and the goal of the task is generally a single target in a specific domain. On the other hand, intelligent outbound call robots sometimes need to imitate human agents to keep the conversation going in order to better accomplish business goals, which is similar to non-task-oriented robots. In summary, we can define the technical classification of intelligent outbound robots as shown in the Fig.\ref{fig:outboundrobots-definition}.

Currently, the industry mainly has the following two ways to implement intelligent outbound robots : (1) flow-based robots and (2) behavior-cloning-based robots. Flow-based robots are mostly finite state-based systems \cite{arora2013dialogue}. This kind of robot relies on pre-defined state nodes and uses the user's input for intent recognition as a transition condition to change the state node to achieve the goal of controlling the flow of the dialogue \cite{mctear1998modelling,yan2017building,lee2008implementation}. Flow-based methods are very popular in the industry due to their excellent ease of implementation and high re-usability. Based on a broad perception, imitating the prior knowledge of experts is more effective and efficient to acquire robotic skills in the real world than finding solutions from scratch \cite{billard2008survey,torabi2018behavioral,florence2021implicit}. Therefore, another kind of solution of intelligent outbound robots is based on behavior-cloning \cite{bain1995framework} methods. Instead of manual dialog flow configuration, behavior-cloning-based robots try to learn and imitate the experience of human agents to choose strategies for responses. But behavior-cloning-based robots also have some shortcomings : (1) due to the randomness of the human strategies, the learning process of the behavior-cloning-based method is complicated, and the models are hard to train. (2) the behavior-cloning-based robot will copy any human behavior in the training set, even if it may be improper behavior. This leads to a bias between the quality of human behavior fitting and the effectiveness of the robot in the real scenario.
In addition, neither the flow-based methods nor the behavior-cloning methods can establish a connection between the dialogue target and the human strategy, which means that they may not be able to select specific optimal strategies for a specific customer.
\section{Policy2target Framework Overview}
The proposed Policy2Target is a framework based on the spoken dialogue system (SDS) for intelligent outbound robots. Generally, an SDS is built by several different modules: automatic speech recognition (ASR), natural language understanding (NLU), dialogue manager (DM), natural language generation (NLG), and text-to-speech (TTS), as shown in Fig.\ref{fig:framework-overview}. The optimization of our framework is mainly for the DM module and the NLG module. 
\subsection{Optimization Target Definition}
We first give a clear definition of the optimization target of the intelligent outbound robots, since previous work on intelligent outbound robots basically draws on question-and-answer robots and task-based robots, but lacks a clear optimization definition. Let $c_{0}$ denote the initial state of a dialogue and $\tau = {c_{0}, a_{0}, c_{1}, a_{1}, ..., c_{T}, a_{ T}}$ is used to represent T rounds of the dialogue interacted with the customer. Let $\pi(a_{t}|c_{t})$ indicate the probability distribution of the behavior made by the robot in the T-th round state $c_{t}$ and let $p_{\pi}(\tau)$ indicate the probability that the robot generates the dialogue $\tau$ based on the $\pi(a_{t}|c_{t})$. Then optimization target of the intelligent outbound robot is find a best $\pi(a_{t}|c_{t})$ to maximum the following formula:
\begin{align}  
R(\pi) = E_{\tau \sim{p_{\pi}(\tau)}}(y|\tau)
\end{align}
where $y$ represents the result after the dialogue ends, such as whether the user repays the payment in the collection scenario.

For the convenience of describing the framework in detail, the following content of this paper will take the collection robot as a specific example of the intelligent outbound robots. It is worth noting that the proposed framework can be used for similar outbound scenarios, e.g., telemarketing.
\subsection{Objectives Applied to Collection Scenarios}
In collection scenarios, the target of the intelligent outbound robots is the repayment of the called debtor.

As illustrated in Fig.\ref{fig:framework-overview}, our framework has some different processes for offline training and online prediction. In offline training, we organize the training data into sequence samples and present them to a dual-way model for multi-task joint learning. During training, the input of a sample includes the collector strategy sequence of all time steps, the user intent sequence of all time steps, and the corresponding user profile features. Differently, when making online predictions, our model will use streaming inference, that is, making predictions for each single time step (each round of the dialogue). When inferring at each time step, we will take all candidate strategies of the current step, the user intent of the current turn, and the user profile features as the inputs of the trained model. Then we will calculate the repayment probabilities of the candidate strategies, and maintain the logical consistency of the context through an implicit context tracker module. Finally, the most effective collection strategy is selected for the current step, and then a script which is the most similar to the customer's words is chosen from the template scripts corresponding to the strategy to reply. The similarity of the script and the customer's words is calculated by a SENTBert \cite{reimers2019sentence}.

It should be noted that a collection strategy here refers to a "human action" or "human strategy", which usually consists of several small strategies. For example, the "identity verification" strategy in Fig.\ref{fig:label} usually consists of a "greeting" strategy and an "asking whether it is the debtor" strategy. A collection strategy in this paper will be referred to as "strategy" or "mixed policy" subsequently. Correspondingly, a small strategy that makes up the "mixed policy" is called "atom strategy" or "atom policy". 

It is assumed that a conversation has $T$ rounds of interaction. In each interaction step $t$, we denote the user's intent obtained by the NLU module as $i_{t}$, and the strategy used by the collector is indicated as $s_{t}$, correspondingly. As mentioned above, the strategy $s_{t}$ may consist of $K$ atomic strategies, which is expressed as $s_{t}=\{s^{t}_{1},...,s^{t}_{K}\}$, where $s^{t}_{k} \in \{0,1\}$.
Given the context information $S = \{s_{1},s_{2},...,s_{t-1}\}$, $I = \{ i_{1},i_{2},...,i_{t-1}\}$ , the current user intent $i_{t}$ and the user profile features $u$, the goal of our model is to predict the repayment probability $\hat{y}_{t}$ and the usage probability $\hat{s}^{t}_{k}$ of the $k$-th atom policy used at $t$-th step :
\begin{align}  
\hat{y}_{t} &= P(y=1|i_{1}, ..., i_{t-1}, s_{1}, ..., s_{t},u) \\
\hat{s}^{t}_{k} &= P(s^{t}_{k}=1|i_{1}, ..., i_{t}, s_{1}, ..., s_{t-1},u)
\end{align}
When the model is used for online inference, we use the idea of maximization to choose from candidate strategies in each step. Let $C$ represent a set of collection strategy candidates used online. For each $s$ in $C$, we
denote the usage score of $s$ as $\hat{s}^{*}$, which is calculated as :
\begin{align}
\hat{s}^{*} = \sum_{k=1}^K s_{k}log\hat{s}_{k} + (1-s_{k})log(1-\hat{s}_{k})
\end{align}
Finally, the choice of the strategy of the $t$-th step is decided as follows:
\begin{align}  
s_{inference,t} &= \mathop{\arg\max}\limits_{s \in C} \hat{y}_{t} * \mathbbm{I}(\hat{s}^{*} > \theta)
\end{align}
where $\hat{y}_{t}$ is related to $s$ as described above, $\theta$ is a threshold parameter and $\mathbbm{I}$ is the indicator function.
\section{Implementation of Policy2Target}
\subsection{Data Preprocessing}
\begin{figure}[htbp]
\centering
\includegraphics[scale=0.3]{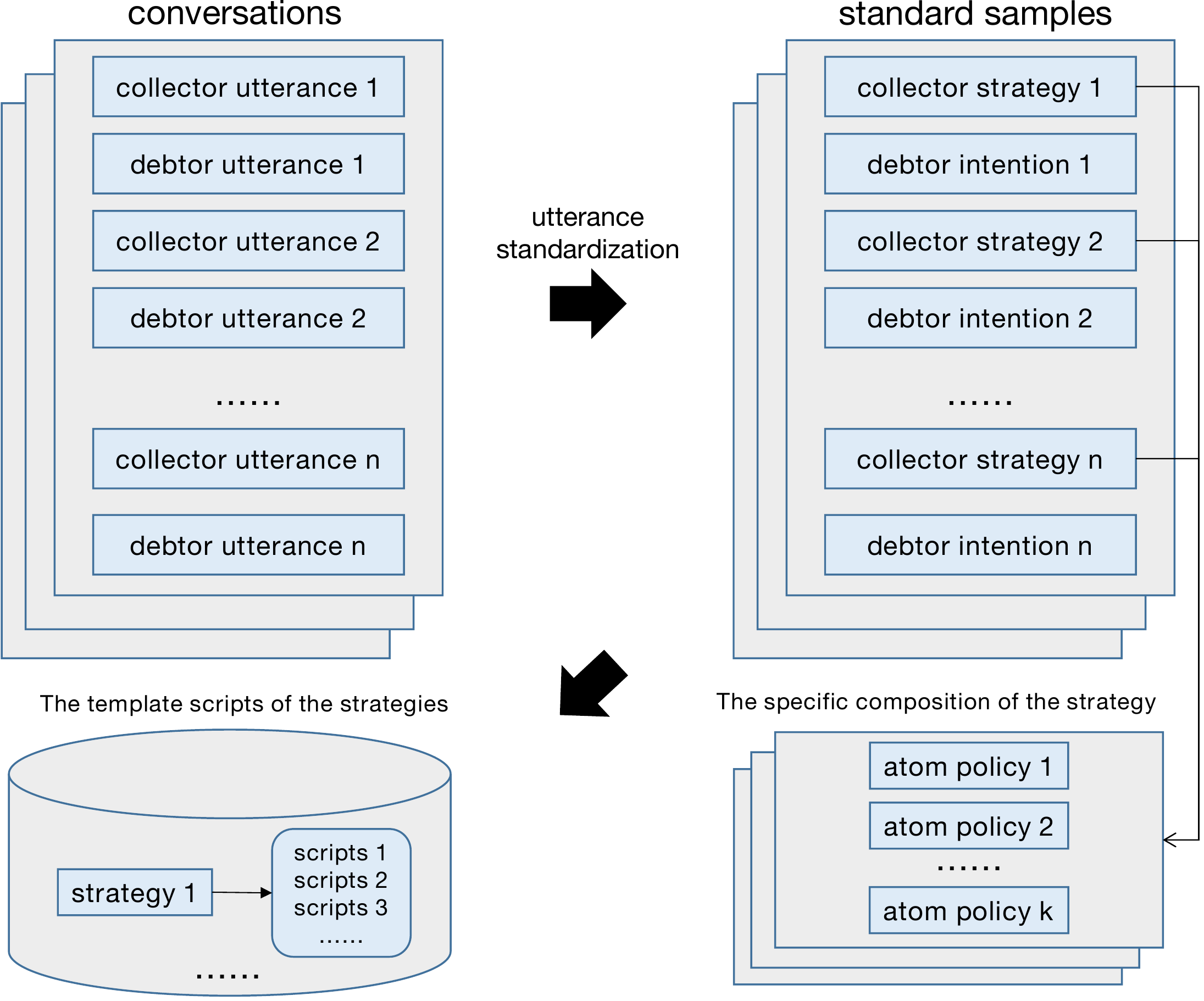}
\caption{Utterance standardization of the conversation.}
\label{fig:data-preprocess}
\end{figure}
The data collected from manual collection calls cannot be used directly for training since the words of the collectors may be complex and changeable, which cannot be exhaustively listed. However, considering that the number of the "collection strategies" of the collectors is limited, we have designed the preprocessing process as shown in the Fig.\ref{fig:data-preprocess}.
After organizing the collection recordings into the text form of multiple rounds of dialogue, we use a multi-label classification model to standardize the collectors' utterances, as depicted in Fig.\ref{fig:data-preprocess}. The collector's words in a round of dialogue are finally standardized as a combination of multiple atomic strategies. In particular, we standardize the user's utterance in each round to the user's intent by using an intent recognition model that is consistent with the online NLU module. 
We also generate the database of template scripts, which is used to return a script to reply after the strategy is selected online. It should be noted that the script semantics of the same strategy is as similar as possible.

The dialogue standardization is necessary since it is beneficial to modeling the strategy use probability more conveniently. In addition, this preprocessing scheme can prevent the model from learning some inexplicable language patterns of the collectors (such as the noise caused by the ASR module, various colloquial expressions of the manual collectors, etc) and avoid deviations in the use of the model. Moreover, we can greatly increase the efficiency of the model calculations to meet online real-time requirements after the standardization.
\subsection{The Attention-based Dual-way Model}
\textbf{User feature processing.} Similar to most CTR models \cite{cheng2016wide,guo2017deepfm,guo2019pal,wang2017deep,zhou2018deep}, the structure of embedding and MLP (Multiple Layer Perception) are used as our user feature processing module. For sparse features, let $u_{1}$ denote the processed dense vectors which are transformed from the sparse user features. Especially, for numerical features, we adopt the autoDis method \cite{guo2020embedding} for automatic discretization and aggregation. And let $u_{2}$ denote the processed dense vectors which are transformed from numerical features. The final processed user feature $u$ is calculated as follows:
\begin{align}
    u = Relu(Concat(u1,u2))
\end{align}
where $Relu(x)$ is the activation function calculated as $Relu(x) = max(0,x)$. 
\\
\textbf{Policy Aggregation.}
Considering that collectors often use many atomic policies when collecting collections, but the importance of different atomic policies to the target is not consistent, we designed an attention mechanism to aggregate the atomic policies in the strategy. 

Given the embedding vector of the $k$-th atomic policy as $P_{k}$ and the embedding vector of the user intent as $I_{t}$ and user feature vector $u$, the first aggregated vector $R'_t$ of the strategy at $t$ step is formulated as follows:
\begin{align}
    e^{t}_{k} &= Tanh(Concat(P_{k},I_{t},u)\boldsymbol{W_{a}}+b_{a})\boldsymbol{W_{b}}+b_{b} \\
    \alpha^{t} &= Softmax(e^t) \\ 
    R'_{t} &= \sum_{k}\alpha^{t}_{k}P_{k}
\end{align}
where $\boldsymbol{W_{a}}$, $b_{a}$, $\boldsymbol{W_{b}}$, $b_{b}$ are attention parameters to be learned.

The second aggregated vector of the strategy indicated by $R"_t$ is inspired by \cite{chen2017sca}, which is tried to be used to adjust the weights of the features automatically, and it is calculated as follows:
\begin{align}
    d^t_{k} & = Concat(P_{k},I_{t},u) \\
    e'^{t}_{k} &= Sigmoid(d^{t}_{k})\boldsymbol{W'_{a}}+b'_{a})\boldsymbol{W'_{b}}+b'_{b} \\
    f^t_{k} &= (d^t_{k} \otimes e'^{t}_{k})\boldsymbol{W'_{c}}+b'_{c} \\
    R''_{t} &= \frac{1}{K}\sum^{K}_{k=1}f^t_{k}
\end{align}
where $\boldsymbol{W'_{a}}$, $b'_{a}$, $\boldsymbol{W'_{b}}$, $b'_{b}$, $\boldsymbol{W'_{c}}$, $b'_{c}$ are parameters to be learned. And $\otimes$ means the element-wise multiplication. $Sigmoid$ is the activation function calculated as $Sigmoid(x)= \frac{1}{1+e^{-x}}$. It is noted that the last dimension of the vector $f^t_{k}$ needs to be adjusted to the same dimension as the embedding size, relying on the parameter $\boldsymbol{W'_{c}}$.
The final aggregated vector $R_t$ of the strategy is summarized as follows:
\begin{align}
    R_{t} &= R'_{t} \oplus R''_{t}
\end{align}
where $\oplus$ means element-wise plus operation.
\\
\textbf{Multi-task Joint Learning.}
We introduce the proposed two-way joint learning model for the Policy2Target framework. As is depicted in Fig.\ref{fig:framework-overview}, the model of Policy2Target consists of two tasks: (1) the repayment probability prediction task and (2) the strategy usage probability prediction task. After obtaining the vector representation $R_{t}$ of the collector’s words, we use $a_{t}=[R_{t},I_{t},u]$ to denote the input for the repayment probability prediction task in one step. Then the inputs of the sequence $A_{T}=\{a_{ 1}, a_{2}, ..., a_{T}\}$ go through the GRU\cite{chung2014empirical} cells and the output can be obtained through a classification layer activated by the sigmoid function. Let $\hat{Y}_{T} = \{\hat{y}_{1}, \hat{ y}_{2}, ..., \hat{y}_{T}\}$ denote the target of the repayment probability prediction task, which is formulated as follows:
\begin{align}  
\hat{Y}_{T} &= Sigmoid((GRU_{1}(A_{T})\boldsymbol{W_{1}}+b_{1})
\end{align}
where $\boldsymbol{W_{1}}$ denotes the training parameters of the classification layer of the task.
The repayment probability prediction task is trained to minimize the cross-entropy loss function:
\begin{align}  
L_{a} &= -\sum^{T}_{t=1}[ylog(\hat{y}_{t}) + (1-y)log(1-\hat{y}_{t})]
\end{align}
where $\hat{y}_{i}$ is the output of the classification layer of the task and $y$ is the binary label indicating the repayment status corresponding to the sample. It is noted that since we don’t know which time step caused the repayment result actually, the labels on all time steps of the sequence corresponding to the same conversation are set to the same. In other words, the labels of each dialogue sequence are the same as the debtor’s repayment status after the call corresponding to the conversation.

Similarly, we use $b_{t}=[R_{t},I_{t},u]$ to represent the input for the strategy usage the probability task. And let $B_{T}=\{b_{1},b_{2},...,b_{T}\}$ represent the target of the strategy usage probability prediction task, and it can be formulated as follows:
\begin{align}  
\hat{S}_{T} &= Sigmoid((GRU_{2}(B_{T})\boldsymbol{W_{2}}+b_{2})
\end{align}
$\boldsymbol{W_{2}}$ also represents the training parameters of the classification layer.
The loss of the strategy usage probability prediction task is summarized as follows correspondingly:
\begin{align}  
L_{b} &= -\sum^{K}_{k=1}\sum^{T}_{t=1}[s^{t}_{k}log\hat{s}^{t}_{k} + (1-s^t_k)log(1-\hat{s}^{t}_{k})]
\end{align}
where $s^{t}_{k}$ indicates whether the $k$-th atomic policy in the $t$-th step appears, which is mentioned before. And $\hat{s}^{t}_{k}$ is the output of the classification layer corresponding to the strategy usage probability prediction task.
For joint training, the final objective function is to minimize the above two loss functions:
\begin{align}
L &= L_{a} + \lambda L_{b}
\end{align}
where $\lambda$ is a hyper-parameter to balance losses.
\\
\textbf{Implicit context tracker.}
\begin{figure}[htbp]
\centering
\includegraphics[scale=0.25]{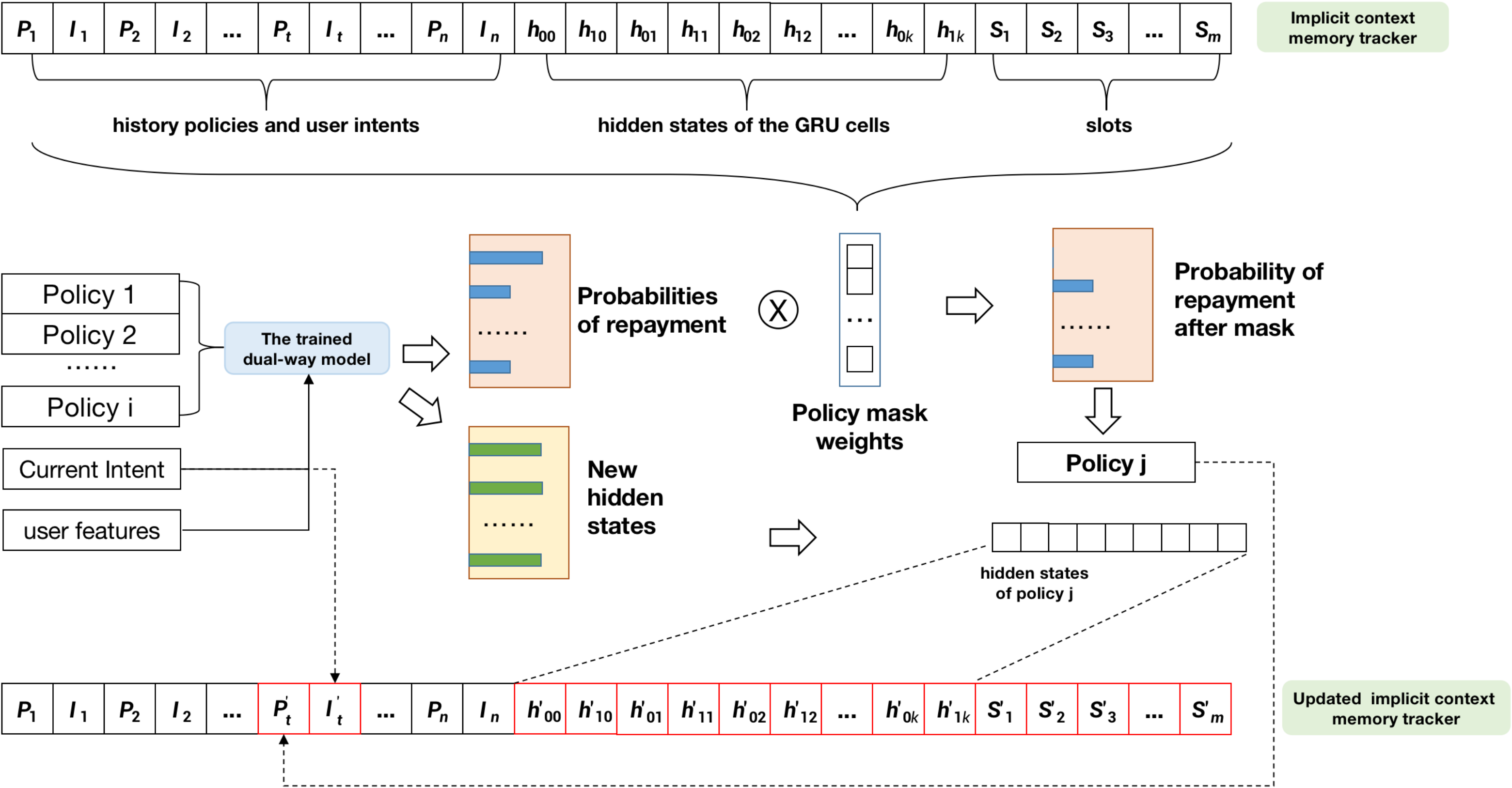}
\caption{Workflow of the memory tracker.}
\label{fig:implicit-tracker}
\end{figure}
In order to successfully deploy the framework online, we also need to consider the following two factors: (1) as described above, the inputs and outputs of the trained model are performed as streaming inference, which is calculated once in each turn of the dialogue. It means that we need to memorize the hidden states of the GRU cells. (2) Limited by actual online scenarios, some strategies may have a strict limit on the number of occurrences. For example, some pressure strategies may be very effective, but they will only be allowed to be triggered a limited number of times. Inspired by \cite{henaff2016tracking}, we designed an implicit context tracker for the proposed framework. As illustrated in Fig.\ref{fig:implicit-tracker}, there are 3 blocks of hidden states acting as a set of read-writeable memories in the implicit context tracker: (1) the first block is responsible for recording history policies and user intents. (2) the second block is used to maintain the vectors of hidden states of GRU cells. (3) the third block is designed to record some special slot values to meet the needs of the scenario, such as the repayment time promised by the debtor. After the model predicts the repayment probabilities of the candidate strategies in each round of dialogue, we generate a vector based on the information provided by the implicit context tracker, and the vector is used to mask the strategies that do not meet the needs of the current round of the scenario. After the strategy is selected, the implicit context tracker will update the values of the corresponding blocks. It is worth mentioning that the implicit context tracker can be configured according to the needs of the scenario.
\section{Experiments} 
In this section, we demonstrate the effectiveness of our Policy2Target framework. The experimental settings and experimental results are described in detail.
\subsection{Dataset \& Experimental Settings}
It is noted the framework is used to learn dialogue skills from the target-oriented human dialogue corpus. Since there is no public dataset that can be applied to our scenario as far as we know, the experiments are conducted on our own dataset.
We constructed a complex dataset\footnote{Data Protection Statement: The Personal Identifiable Information(PII) in this research has been cleaned up to ensure that no privacy leakage will be involved.
} of multi-round of dialogue which is transcribed by ASR from a large number of online human-human telephone calls between collectors and debtors. The word accuracy rate of ASR is about 89\%. 
For the multi-label strategy classification model in data preprocessing, we use BERT-Large \cite{devlin2018bert} as our preprocessing classification model. The average accuracy of each label of the preprocessing classification model is about 71\%. For the dual-way model, the longest round of a conversation is set at 50, and the upper limit of the number of atomic policies included in a strategy is set to 10.
\subsection{Evaluation}
\textbf{Offline evaluation.} We use AUC (Area Under ROC) as one of our offline evaluation metrics. Furthermore, considering that the user profile is highly related to the scene target, we propose Weighted Bins AUC (WBAUC) as another evaluation metric to ensure our model can learn the strategy selection bias instead of overfitting user features. Firstly, we divide the dialogues in the test dataset into multiple bins based on their user profile information. Next, we calculate the AUC score of each bin, and obtain the final score by weighted average operation, which is formulated as follows:
\begin{align}
WBAUC=\frac{\sum_{i=1}^{N}bin_{i}AUC_{i}}{\sum_{i=1}^{N}bin_{i}}
\end{align}
where N denotes the number of bins divided, $bin_{i}$ denotes the number of samples in the $i$-th bin, and $AUC_{i}$ is the AUC of the $i$-th bin. The WBAUC measures the relevance ranking quality at every bin, reducing the impact of the user feature bias. It is noted that we use the repayment probability predicted for the last step as the prediction for each dialogue to calculate the metrics.
Since the flow-based method and the behavior-cloning-based (BC-based) method are not designed to predict the probability of the repayment, we regard their prediction of repayment as random float numbers between 0.5 and 1 to calculate AUC and WBAUC. And the following methods are designed and compared offline:
\\
(1) We use a binary classification \cite{ke2017lightgbm} model with only user features as a baseline, which is denoted as "BCU" model. \\
(2) The proposed Policy2Target model, which is denoted as "P2T" model.\\
(3) The P2T model without user features. \\
(4) The multi-task in the P2T model is replaced with a single task with only one-way GRU. \\
(5) The P2T model without the attention-based policy aggregation module.\\
The details of the AUC and WBAUC are shown in Table \ref{table:evaluation-target-prediction}.
\begin{table}[htbp]
\centering
\caption{Evaluation on Target Effect Prediction}
\begin{tabular}{@{}c|c|c@{}}
\toprule
Method           &AUC      & WBAUC \\ \midrule
Flow-based \& BC-based model & 0.4989 & 0.4978 \\
BCU model & 0.7554 & 0.5116 \\ 
P2T-without-user-features & 0.7488 & 0.6811 \\
P2T-single-way-task & 0.8139 & 0.6900 \\
P2T-w/o-attention-aggregation & 0.8126 & 0.6932   \\
\textbf{P2T model}   & \textbf{0.8114}  & \textbf{0.6972}   \\
\bottomrule
\end{tabular}
\label{table:evaluation-target-prediction}
\end{table}
 As illustrated in Table \ref{table:evaluation-target-prediction}, the baseline method has a huge gap between AUC and WBAUC, which indicates that the repayment prediction target is highly correlated with user information. And the bins we divided are qualified since the similarity of user features within each bin is high enough to get a relatively small WBAUC. Compared with the baseline that only uses user profile features, the strategy selection model brings a huge improvement for the prediction of the repayment on WBAUC. At the same time, it can be observed that the gap between the AUC and the WBAUC of the Policy2Target model is relatively small, which can prove that the model reduces the dependence of target prediction on the user feature bias. The comparison between P2T-without-user-features and the BCU model shows that the repayment prediction ability of the selection of strategies is not as good as user features for the AUC, but the P2T-without-user-features performs better on the WBAUC, which proves that the structure of the P2T model can learn the strategy selection bias well. And the complete P2T gains 6.26\% on AUC and gains 1.61\%  on WBAUC compared with P2T-without-user-features, achieving consistency, which shows the P2T model can alleviate the bias brought by user features while making full use of the features. Besides, from the results of the other experiments, we can summarize that: (1) the structure of the Policy2Target model mainly benefits from 
the dual-task joint learning. (2) The improvement is also from the policy aggregation, though not as significant as the former. Both modules help reduce the impact of the user feature bias since there will have a performance improvement on AUC but a degradation on WBAUC after removing them.
\\
\textbf{Simulator evaluation.} In order to evaluate the dialogue quality of the dialogue systems, we also develop a user simulator and employ the simulator to converse with the policy2target robot, the behavior-cloning robot, and the flow-based robot respectively. The flow-based robot implemented here is similar to \cite{yan2017building} and the behavior-cloning robot adopted is similar to \cite{wang2020two}. Both of the compared methods have been highly optimized for the scenario and are strong baselines.

We use the average number of the dialogue rounds and the dialogue diversity as metrics for the simulator evaluation. The number of the dialogue rounds means the overall conversation turns between the robot and customers. For the dialogue diversity, we calculate the proportion of distinct dialogue paths, which is formulated as follows:
\begin{align}
    Diversity = \frac{Deduplicate(dialogue\ paths)}{dialogue\ paths} 
\end{align}
where a dialogue path of a conversation means the robot utterances from the first turn to the last turn in the conversation. 

It can be seen from Table \ref{table:average-dialogue-rounds-diversity} that our proposed framework performs better than both behavior-cloning-based (BC-based) robots and flow-based robots. Since our flow-based method has been optimized by a large number of experts, both of the two model-based methods (BC-based and Policy2Target) have a relatively small advantage in the average dialogue round compared with the Flow-based method. But for dialogue diversity, both of the model-based methods have a huge advantage compared with the flow-based method, indicating that the model-based methods are more complicated and more “human-like”. Especially, compared with the behavior-cloning-based method, the Policy2Target method has about 17.4\% gain on dialogue diversity, indicating that the Policy2Target method can make better use of user features to select collection strategies.
\begin{table}[htbp]
\caption{The average dialogue round and dialogue diversity of simulator evaluation.}
\centering
\begin{tabular}{@{}c|c|c@{}}
\toprule
Method         & Dialogue Rounds & Dialogue Diversity \\ \midrule
BC-based           & 4.48             & 50.8\%               \\
Flow-based     & 4.07             & 28.5\%               \\
\textbf{Policy2Target} & 4.89             & 68.2\%               \\
\bottomrule
\end{tabular}
\label{table:average-dialogue-rounds-diversity}
\end{table}
\\
\textbf{Online evaluation.}
To evaluate the performance of Policy2Target in real scenarios, we conducted an online A/B test in the debt collection system from 2021-07-03 to 2021-09-30. The debtors to be called are randomly sorted firstly. For the control groups, 10\% of debtors are called by the optimized flow-based robots, and another 10\% of debtors are called by behavior-cloning-based robots. For the experimental group, 10\% of debtors are called by Policy2Target robots. A/B test shows that the Policy2Target has improved the rate of repayment by 1.60\% compared to the flow-based method and 0.85\% compared to the behavior-cloning-based method. For now, Policy2Target has been deployed online and serves the main traffic, which helps business revenue to achieve significant growth. The online result of the average of the dialogue rounds and the diversity is shown as \ref{table:average-dialogue-rounds-diversity-on}, and the conclusions obtained are consistent with the result of the simulator evaluation.

\begin{table}[htbp]
\caption{The average dialogue round and dialogue diversity online.}
\centering
\begin{tabular}{@{}c|c|c@{}}
\toprule
Method         & Dialogue Rounds & Dialogue Diversity \\ \midrule
BC-based           & 4.51             & 50.7\%               \\
Flow-based     & 4.05             & 26.4\%               \\ 
\textbf{Policy2Target} & 4.91             & 68.5\%               \\
\bottomrule
\end{tabular}
\label{table:average-dialogue-rounds-diversity-on}
\end{table}

\section{Conclusion} 
In this work, we demonstrate the previous solutions of intelligent outbound robots in the industry. A novel target-driven framework named Policy2Target is designed to establish the mapping of the dialogue strategy to the dialogue target accomplishment probability. And an implementation for a debt collection robot is described in detail as an example of the proposed framework. A new evaluation metric called WBAUC is proposed for measuring the performance of the model, solving the bias problem caused by user features. The experiments show that the proposed Policy2Target outperforms the compared methods. In addition, an online A/B test is also conducted to demonstrate that Policy2Target has a significant improvement over previous solutions. For the moment, the Policy2Target robot is deployed in an intelligent debt collection system in a FinTech company, serving the main traffic.



\bibliographystyle{refer_format}
\bibliography{sample_base}

\begin{thebibliography}{30}


\ifx \showCODEN    \undefined \def \showCODEN     #1{\unskip}     \fi
\ifx \showDOI      \undefined \def \showDOI       #1{#1}\fi
\ifx \showISBNx    \undefined \def \showISBNx     #1{\unskip}     \fi
\ifx \showISBNxiii \undefined \def \showISBNxiii  #1{\unskip}     \fi
\ifx \showISSN     \undefined \def \showISSN      #1{\unskip}     \fi
\ifx \showLCCN     \undefined \def \showLCCN      #1{\unskip}     \fi
\ifx \shownote     \undefined \def \shownote      #1{#1}          \fi
\ifx \showarticletitle \undefined \def \showarticletitle #1{#1}   \fi
\ifx \showURL      \undefined \def \showURL       {\relax}        \fi
\providecommand\bibfield[2]{#2}
\providecommand\bibinfo[2]{#2}
\providecommand\natexlab[1]{#1}
\providecommand\showeprint[2][]{arXiv:#2}

\bibitem[Arora et~al\mbox{.}(2013)]%
        {arora2013dialogue}
\bibfield{author}{\bibinfo{person}{Suket Arora}, \bibinfo{person}{Kamaljeet
  Batra}, {and} \bibinfo{person}{Sarabjit Singh}.}
  \bibinfo{year}{2013}\natexlab{}.
\newblock \showarticletitle{Dialogue system: A brief review}.
\newblock \bibinfo{journal}{\emph{arXiv preprint arXiv:1306.4134}}
  (\bibinfo{year}{2013}).
\newblock


\bibitem[Bain and Sammut(1995)]%
        {bain1995framework}
\bibfield{author}{\bibinfo{person}{Michael Bain} {and} \bibinfo{person}{Claude
  Sammut}.} \bibinfo{year}{1995}\natexlab{}.
\newblock \showarticletitle{A Framework for Behavioural Cloning.}. In
  \bibinfo{booktitle}{\emph{Machine Intelligence 15}}.
  \bibinfo{pages}{103--129}.
\newblock


\bibitem[Balakrishnan et~al\mbox{.}(2019)]%
        {balakrishnan2019constrained}
\bibfield{author}{\bibinfo{person}{Anusha Balakrishnan},
  \bibinfo{person}{Jinfeng Rao}, \bibinfo{person}{Kartikeya Upasani},
  \bibinfo{person}{Michael White}, {and} \bibinfo{person}{Rajen Subba}.}
  \bibinfo{year}{2019}\natexlab{}.
\newblock \showarticletitle{Constrained decoding for neural NLG from
  compositional representations in task-oriented dialogue}.
\newblock \bibinfo{journal}{\emph{arXiv preprint arXiv:1906.07220}}
  (\bibinfo{year}{2019}).
\newblock


\bibitem[Billard et~al\mbox{.}(2008)]%
        {billard2008survey}
\bibfield{author}{\bibinfo{person}{Aude Billard}, \bibinfo{person}{Sylvain
  Calinon}, \bibinfo{person}{Ruediger Dillmann}, {and} \bibinfo{person}{Stefan
  Schaal}.} \bibinfo{year}{2008}\natexlab{}.
\newblock \bibinfo{booktitle}{\emph{Survey: Robot programming by
  demonstration}}.
\newblock \bibinfo{type}{{T}echnical {R}eport}.
  \bibinfo{institution}{Springrer}.
\newblock


\bibitem[Chen et~al\mbox{.}(2017a)]%
        {chen2017survey}
\bibfield{author}{\bibinfo{person}{Hongshen Chen}, \bibinfo{person}{Xiaorui
  Liu}, \bibinfo{person}{Dawei Yin}, {and} \bibinfo{person}{Jiliang Tang}.}
  \bibinfo{year}{2017}\natexlab{a}.
\newblock \showarticletitle{A survey on dialogue systems: Recent advances and
  new frontiers}.
\newblock \bibinfo{journal}{\emph{Acm Sigkdd Explorations Newsletter}}
  \bibinfo{volume}{19}, \bibinfo{number}{2} (\bibinfo{year}{2017}),
  \bibinfo{pages}{25--35}.
\newblock


\bibitem[Chen et~al\mbox{.}(2017b)]%
        {chen2017sca}
\bibfield{author}{\bibinfo{person}{Long Chen}, \bibinfo{person}{Hanwang Zhang},
  \bibinfo{person}{Jun Xiao}, \bibinfo{person}{Liqiang Nie},
  \bibinfo{person}{Jian Shao}, \bibinfo{person}{Wei Liu}, {and}
  \bibinfo{person}{Tat-Seng Chua}.} \bibinfo{year}{2017}\natexlab{b}.
\newblock \showarticletitle{Sca-cnn: Spatial and channel-wise attention in
  convolutional networks for image captioning}. In
  \bibinfo{booktitle}{\emph{Proceedings of the IEEE conference on computer
  vision and pattern recognition}}. \bibinfo{pages}{5659--5667}.
\newblock


\bibitem[Cheng et~al\mbox{.}(2016)]%
        {cheng2016wide}
\bibfield{author}{\bibinfo{person}{Heng-Tze Cheng}, \bibinfo{person}{Levent
  Koc}, \bibinfo{person}{Jeremiah Harmsen}, \bibinfo{person}{Tal Shaked},
  \bibinfo{person}{Tushar Chandra}, \bibinfo{person}{Hrishi Aradhye},
  \bibinfo{person}{Glen Anderson}, \bibinfo{person}{Greg Corrado},
  \bibinfo{person}{Wei Chai}, \bibinfo{person}{Mustafa Ispir}, {et~al\mbox{.}}}
  \bibinfo{year}{2016}\natexlab{}.
\newblock \showarticletitle{Wide \& deep learning for recommender systems}. In
  \bibinfo{booktitle}{\emph{Proceedings of the 1st workshop on deep learning
  for recommender systems}}. \bibinfo{pages}{7--10}.
\newblock


\bibitem[Chung et~al\mbox{.}(2014)]%
        {chung2014empirical}
\bibfield{author}{\bibinfo{person}{Junyoung Chung}, \bibinfo{person}{Caglar
  Gulcehre}, \bibinfo{person}{KyungHyun Cho}, {and} \bibinfo{person}{Yoshua
  Bengio}.} \bibinfo{year}{2014}\natexlab{}.
\newblock \showarticletitle{Empirical evaluation of gated recurrent neural
  networks on sequence modeling}.
\newblock \bibinfo{journal}{\emph{arXiv preprint arXiv:1412.3555}}
  (\bibinfo{year}{2014}).
\newblock


\bibitem[Dai et~al\mbox{.}(2020)]%
        {dai2020learning}
\bibfield{author}{\bibinfo{person}{Yinpei Dai}, \bibinfo{person}{Hangyu Li},
  \bibinfo{person}{Chengguang Tang}, \bibinfo{person}{Yongbin Li},
  \bibinfo{person}{Jian Sun}, {and} \bibinfo{person}{Xiaodan Zhu}.}
  \bibinfo{year}{2020}\natexlab{}.
\newblock \showarticletitle{Learning low-resource end-to-end goal-oriented
  dialog for fast and reliable system deployment}. In
  \bibinfo{booktitle}{\emph{Proceedings of the 58th Annual Meeting of the
  Association for Computational Linguistics}}. \bibinfo{pages}{609--618}.
\newblock


\bibitem[Devlin et~al\mbox{.}(2018)]%
        {devlin2018bert}
\bibfield{author}{\bibinfo{person}{Jacob Devlin}, \bibinfo{person}{Ming-Wei
  Chang}, \bibinfo{person}{Kenton Lee}, {and} \bibinfo{person}{Kristina
  Toutanova}.} \bibinfo{year}{2018}\natexlab{}.
\newblock \showarticletitle{Bert: Pre-training of deep bidirectional
  transformers for language understanding}.
\newblock \bibinfo{journal}{\emph{arXiv preprint arXiv:1810.04805}}
  (\bibinfo{year}{2018}).
\newblock


\bibitem[Florence et~al\mbox{.}(2021)]%
        {florence2021implicit}
\bibfield{author}{\bibinfo{person}{Pete Florence}, \bibinfo{person}{Corey
  Lynch}, \bibinfo{person}{Andy Zeng}, \bibinfo{person}{Oscar Ramirez},
  \bibinfo{person}{Ayzaan Wahid}, \bibinfo{person}{Laura Downs},
  \bibinfo{person}{Adrian Wong}, \bibinfo{person}{Johnny Lee},
  \bibinfo{person}{Igor Mordatch}, {and} \bibinfo{person}{Jonathan Tompson}.}
  \bibinfo{year}{2021}\natexlab{}.
\newblock \showarticletitle{Implicit behavioral cloning}.
\newblock \bibinfo{journal}{\emph{arXiv preprint arXiv:2109.00137}}
  (\bibinfo{year}{2021}).
\newblock


\bibitem[Guo et~al\mbox{.}(2020)]%
        {guo2020embedding}
\bibfield{author}{\bibinfo{person}{Huifeng Guo}, \bibinfo{person}{Bo Chen},
  \bibinfo{person}{Ruiming Tang}, \bibinfo{person}{Weinan Zhang},
  \bibinfo{person}{Zhenguo Li}, {and} \bibinfo{person}{Xiuqiang He}.}
  \bibinfo{year}{2020}\natexlab{}.
\newblock \showarticletitle{An Embedding Learning Framework for Numerical
  Features in CTR Prediction}.
\newblock \bibinfo{journal}{\emph{arXiv preprint arXiv:2012.08986}}
  (\bibinfo{year}{2020}).
\newblock


\bibitem[Guo et~al\mbox{.}(2017)]%
        {guo2017deepfm}
\bibfield{author}{\bibinfo{person}{Huifeng Guo}, \bibinfo{person}{Ruiming
  Tang}, \bibinfo{person}{Yunming Ye}, \bibinfo{person}{Zhenguo Li}, {and}
  \bibinfo{person}{Xiuqiang He}.} \bibinfo{year}{2017}\natexlab{}.
\newblock \showarticletitle{DeepFM: a factorization-machine based neural
  network for CTR prediction}.
\newblock \bibinfo{journal}{\emph{arXiv preprint arXiv:1703.04247}}
  (\bibinfo{year}{2017}).
\newblock


\bibitem[Guo et~al\mbox{.}(2019)]%
        {guo2019pal}
\bibfield{author}{\bibinfo{person}{Huifeng Guo}, \bibinfo{person}{Jinkai Yu},
  \bibinfo{person}{Qing Liu}, \bibinfo{person}{Ruiming Tang}, {and}
  \bibinfo{person}{Yuzhou Zhang}.} \bibinfo{year}{2019}\natexlab{}.
\newblock \showarticletitle{PAL: a position-bias aware learning framework for
  CTR prediction in live recommender systems}. In
  \bibinfo{booktitle}{\emph{Proceedings of the 13th ACM Conference on
  Recommender Systems}}. \bibinfo{pages}{452--456}.
\newblock


\bibitem[Henaff et~al\mbox{.}(2016)]%
        {henaff2016tracking}
\bibfield{author}{\bibinfo{person}{Mikael Henaff}, \bibinfo{person}{Jason
  Weston}, \bibinfo{person}{Arthur Szlam}, \bibinfo{person}{Antoine Bordes},
  {and} \bibinfo{person}{Yann LeCun}.} \bibinfo{year}{2016}\natexlab{}.
\newblock \showarticletitle{Tracking the world state with recurrent entity
  networks}.
\newblock \bibinfo{journal}{\emph{arXiv preprint arXiv:1612.03969}}
  (\bibinfo{year}{2016}).
\newblock


\bibitem[Henderson et~al\mbox{.}(2014)]%
        {henderson2014second}
\bibfield{author}{\bibinfo{person}{Matthew Henderson}, \bibinfo{person}{Blaise
  Thomson}, {and} \bibinfo{person}{Jason~D Williams}.}
  \bibinfo{year}{2014}\natexlab{}.
\newblock \showarticletitle{The second dialog state tracking challenge}. In
  \bibinfo{booktitle}{\emph{Proceedings of the 15th annual meeting of the
  special interest group on discourse and dialogue (SIGDIAL)}}.
  \bibinfo{pages}{263--272}.
\newblock


\bibitem[Hu et~al\mbox{.}(2020)]%
        {hu2020sas}
\bibfield{author}{\bibinfo{person}{Jiaying Hu}, \bibinfo{person}{Yan Yang},
  \bibinfo{person}{Chencai Chen}, \bibinfo{person}{Liang He}, {and}
  \bibinfo{person}{Zhou Yu}.} \bibinfo{year}{2020}\natexlab{}.
\newblock \showarticletitle{SAS: Dialogue state tracking via slot attention and
  slot information sharing}. In \bibinfo{booktitle}{\emph{Proceedings of the
  58th Annual Meeting of the Association for Computational Linguistics}}.
  \bibinfo{pages}{6366--6375}.
\newblock


\bibitem[Ke et~al\mbox{.}(2017)]%
        {ke2017lightgbm}
\bibfield{author}{\bibinfo{person}{Guolin Ke}, \bibinfo{person}{Qi Meng},
  \bibinfo{person}{Thomas Finley}, \bibinfo{person}{Taifeng Wang},
  \bibinfo{person}{Wei Chen}, \bibinfo{person}{Weidong Ma},
  \bibinfo{person}{Qiwei Ye}, {and} \bibinfo{person}{Tie-Yan Liu}.}
  \bibinfo{year}{2017}\natexlab{}.
\newblock \showarticletitle{Lightgbm: A highly efficient gradient boosting
  decision tree}.
\newblock \bibinfo{journal}{\emph{Advances in neural information processing
  systems}}  \bibinfo{volume}{30} (\bibinfo{year}{2017}),
  \bibinfo{pages}{3146--3154}.
\newblock


\bibitem[Kim et~al\mbox{.}(2018)]%
        {kim2018two}
\bibfield{author}{\bibinfo{person}{A Kim}, \bibinfo{person}{Hyun-Je Song},
  \bibinfo{person}{Seong-Bae Park}, {et~al\mbox{.}}}
  \bibinfo{year}{2018}\natexlab{}.
\newblock \showarticletitle{A two-step neural dialog state tracker for
  task-oriented dialog processing}.
\newblock \bibinfo{journal}{\emph{Computational intelligence and neuroscience}}
   \bibinfo{volume}{2018} (\bibinfo{year}{2018}).
\newblock


\bibitem[Lee et~al\mbox{.}(2008)]%
        {lee2008implementation}
\bibfield{author}{\bibinfo{person}{Changyoon Lee}, \bibinfo{person}{You-Sung
  Cha}, {and} \bibinfo{person}{Tae-Yong Kuc}.} \bibinfo{year}{2008}\natexlab{}.
\newblock \showarticletitle{Implementation of dialogue system for intelligent
  service robots}. In \bibinfo{booktitle}{\emph{2008 International Conference
  on Control, Automation and Systems}}. IEEE, \bibinfo{pages}{2038--2042}.
\newblock


\bibitem[McTear(1998)]%
        {mctear1998modelling}
\bibfield{author}{\bibinfo{person}{Michael~F McTear}.}
  \bibinfo{year}{1998}\natexlab{}.
\newblock \showarticletitle{Modelling spoken dialogues with state transition
  diagrams: experiences with the CSLU toolkit}.
\newblock \bibinfo{journal}{\emph{development}} \bibinfo{volume}{5},
  \bibinfo{number}{7} (\bibinfo{year}{1998}).
\newblock


\bibitem[Ni et~al\mbox{.}(2021)]%
        {ni2021recent}
\bibfield{author}{\bibinfo{person}{Jinjie Ni}, \bibinfo{person}{Tom Young},
  \bibinfo{person}{Vlad Pandelea}, \bibinfo{person}{Fuzhao Xue},
  \bibinfo{person}{Vinay Adiga}, {and} \bibinfo{person}{Erik Cambria}.}
  \bibinfo{year}{2021}\natexlab{}.
\newblock \showarticletitle{Recent advances in deep learning based dialogue
  systems: A systematic survey}.
\newblock \bibinfo{journal}{\emph{arXiv preprint arXiv:2105.04387}}
  (\bibinfo{year}{2021}).
\newblock


\bibitem[Reimers and Gurevych(2019)]%
        {reimers2019sentence}
\bibfield{author}{\bibinfo{person}{Nils Reimers} {and} \bibinfo{person}{Iryna
  Gurevych}.} \bibinfo{year}{2019}\natexlab{}.
\newblock \showarticletitle{Sentence-bert: Sentence embeddings using siamese
  bert-networks}.
\newblock \bibinfo{journal}{\emph{arXiv preprint arXiv:1908.10084}}
  (\bibinfo{year}{2019}).
\newblock


\bibitem[Torabi et~al\mbox{.}(2018)]%
        {torabi2018behavioral}
\bibfield{author}{\bibinfo{person}{Faraz Torabi}, \bibinfo{person}{Garrett
  Warnell}, {and} \bibinfo{person}{Peter Stone}.}
  \bibinfo{year}{2018}\natexlab{}.
\newblock \showarticletitle{Behavioral cloning from observation}.
\newblock \bibinfo{journal}{\emph{arXiv preprint arXiv:1805.01954}}
  (\bibinfo{year}{2018}).
\newblock


\bibitem[Wang et~al\mbox{.}(2017)]%
        {wang2017deep}
\bibfield{author}{\bibinfo{person}{Ruoxi Wang}, \bibinfo{person}{Bin Fu},
  \bibinfo{person}{Gang Fu}, {and} \bibinfo{person}{Mingliang Wang}.}
  \bibinfo{year}{2017}\natexlab{}.
\newblock \showarticletitle{Deep \& cross network for ad click predictions}.
\newblock In \bibinfo{booktitle}{\emph{Proceedings of the ADKDD'17}}.
  \bibinfo{pages}{1--7}.
\newblock


\bibitem[Wang et~al\mbox{.}(2019)]%
        {wang2019incremental}
\bibfield{author}{\bibinfo{person}{Weikang Wang}, \bibinfo{person}{Jiajun
  Zhang}, \bibinfo{person}{Qian Li}, \bibinfo{person}{Mei-Yuh Hwang},
  \bibinfo{person}{Chengqing Zong}, {and} \bibinfo{person}{Zhifei Li}.}
  \bibinfo{year}{2019}\natexlab{}.
\newblock \showarticletitle{Incremental learning from scratch for task-oriented
  dialogue systems}.
\newblock \bibinfo{journal}{\emph{arXiv preprint arXiv:1906.04991}}
  (\bibinfo{year}{2019}).
\newblock


\bibitem[Wang et~al\mbox{.}(2020)]%
        {wang2020two}
\bibfield{author}{\bibinfo{person}{Zihao Wang}, \bibinfo{person}{Jia Liu},
  \bibinfo{person}{Hengbin Cui}, \bibinfo{person}{Chunxiang Jin},
  \bibinfo{person}{Minghui Yang}, \bibinfo{person}{Yafang Wang},
  \bibinfo{person}{Xiaolong Li}, {and} \bibinfo{person}{Renxin Mao}.}
  \bibinfo{year}{2020}\natexlab{}.
\newblock \showarticletitle{Two-stage Behavior Cloning for Spoken Dialogue
  System in Debt Collection.}. In \bibinfo{booktitle}{\emph{IJCAI}}.
  \bibinfo{pages}{4633--4639}.
\newblock


\bibitem[Wang et~al\mbox{.}(2021)]%
        {wang2021ifdds}
\bibfield{author}{\bibinfo{person}{Zihao Wang}, \bibinfo{person}{Minghui Yang},
  \bibinfo{person}{Chunxiang Jin}, \bibinfo{person}{Jia Liu},
  \bibinfo{person}{Zujie Wen}, \bibinfo{person}{Saishuai Liu}, {and}
  \bibinfo{person}{Zhe Zhang}.} \bibinfo{year}{2021}\natexlab{}.
\newblock \showarticletitle{IFDDS: An Anti-fraud Outbound Robot}. In
  \bibinfo{booktitle}{\emph{Proceedings of the AAAI Conference on Artificial
  Intelligence}}, Vol.~\bibinfo{volume}{35}. \bibinfo{pages}{16117--16119}.
\newblock


\bibitem[Yan et~al\mbox{.}(2017)]%
        {yan2017building}
\bibfield{author}{\bibinfo{person}{Zhao Yan}, \bibinfo{person}{Nan Duan},
  \bibinfo{person}{Peng Chen}, \bibinfo{person}{Ming Zhou},
  \bibinfo{person}{Jianshe Zhou}, {and} \bibinfo{person}{Zhoujun Li}.}
  \bibinfo{year}{2017}\natexlab{}.
\newblock \showarticletitle{Building task-oriented dialogue systems for online
  shopping}. In \bibinfo{booktitle}{\emph{Thirty-First AAAI Conference on
  Artificial Intelligence}}.
\newblock


\bibitem[Zhou et~al\mbox{.}(2018)]%
        {zhou2018deep}
\bibfield{author}{\bibinfo{person}{Guorui Zhou}, \bibinfo{person}{Xiaoqiang
  Zhu}, \bibinfo{person}{Chenru Song}, \bibinfo{person}{Ying Fan},
  \bibinfo{person}{Han Zhu}, \bibinfo{person}{Xiao Ma},
  \bibinfo{person}{Yanghui Yan}, \bibinfo{person}{Junqi Jin},
  \bibinfo{person}{Han Li}, {and} \bibinfo{person}{Kun Gai}.}
  \bibinfo{year}{2018}\natexlab{}.
\newblock \showarticletitle{Deep interest network for click-through rate
  prediction}. In \bibinfo{booktitle}{\emph{Proceedings of the 24th ACM SIGKDD
  International Conference on Knowledge Discovery \& Data Mining}}.
  \bibinfo{pages}{1059--1068}.
\newblock


\end{thebibliography}










\end{document}